\documentclass[runningheads]{llncs}

\usepackage{eccv}

\usepackage{eccvabbrv}

\usepackage{graphicx}
\usepackage{booktabs}

\usepackage[accsupp]{axessibility}  

\usepackage{hyperref}

\usepackage{orcidlink}

\begin{document}

\title{FreeFly-Thinking : Aligning Chain-of-Thought Reasoning with Continuous UAV Navigation}

\titlerunning{FreeFly-Thinking}

\author{Jiaxu Zhou\inst{1} \and
Shaobo Wang\inst{1,2}
\and
Zhiyuan Yang\inst{3}
\and
Zhenjun Yu\inst{1}
\and
Tao Li\inst{1}}

\authorrunning{J. Zhou et al.}

\institute{SJTU Paris Elite Institute of Technology, Shanghai Jiao Tong University, China \\
\email{\{1729722474-qq.com, shaobo.wang, jeffson-yu , taoli\}@sjtu.edu.cn} \and
School of Physics and Astronomy, Shanghai Jiao Tong University, China \and
Tongji University, China \\
\email{2253350@tongji.edu.cn}}

\maketitle

\begin{abstract}
Vision-Language Navigation(VLN) aims to enable agents to understand natural language instructions and carry out appropriate navigation actions in real-world environments. Most work focuses on indoor settings, with little research in complex outdoor scenes. Current UAV Vision-and-Language Navigation models typically act as black boxes without explicit reasoning. We introduce FreeFly-thinking, an end-to-end VLN framework that converts the UAV agent’s egocentric images and language instructions into a series of actions, inspired by environment of urban architecture proposed by OpenFly. We first construct a UAV dataset for navigation task, and then performing natural language chain of thought. We adopt a two-stage training strategy: Supervised fine-tuning(SFT) and Reinforcement fine-tuning(RFT). Experiments on unseen test demonstrate a strong performance, presenting robustness and efficiency in UAV navigation issue.
  \keywords{Vision-Language Navigation \and UAV \and Cot Reasoning}
\end{abstract}

\section{Introduction}
\label{sec:intro}

Embodied AI has emerged as a pivotal frontier in artificial intelligence, aiming to bridge the gap between high-level semantic understanding and low-level robotic execution by enabling autonomous agents to perceive, reason, and navigate within complex, real-world environments\cite{habitat, survey1, survey2}. In recent years, Vision-and-Language Navigation(VLN)\cite{vln, navgpt2} has sparked widespread research interest, which stands as a fundamental challenge in the pursuit of Embodied AI. Specifically, this task requires the agent to parse complex natural language instructions and dynamically correlate them with visual observations and environmental information to execute a series of navigation maneuvers toward the target destination. Existing research predominantly focuses on indoor environments (e.g., R2R\cite{r2r}, RxR\cite{rxr}, VLN-CE\cite{vlnce}) and ground-based navigation, leaving outdoor and Unmanned Aerial Vehicle(UAV) scenarios comparatively under-explored. Given the widespread adoption of UAVs in fields such as environmental sensing and autonomous logistics \cite{survey3}, there has been a significant surge in the requirement for robust systems that can accurately translate human intentions into specific navigation actions.

\begin{figure}[tb]
  \centering
  \includegraphics[height=6.5cm]{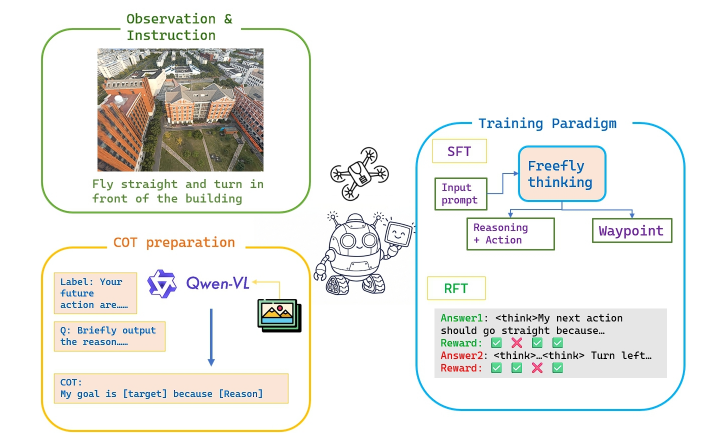}
  \caption{Overview for our work, illustrating the complete pipeline from automated Chain-of-Thought (CoT) data preparation based on visual observations and instructions, to the two-stage training paradigm (SFT and RFT) that optimizes the dual-head Freefly-thinking model for UAV navigation.}
  \label{fig:overview}
\end{figure}

Current UAV\cite{uavvln, fly0} VLN methods predominantly rely on black-box architectures mapping multimodal inputs directly to discrete actions. This paradigm lacks both interpretability and physical grounding, failing to bridge the semantic-to-control gap between high-level instructions and low-level kinematics. This limitation arises because effective UAV navigation requires a cohesive Image-Instruction-Control logic chain, where reasoning serves as the essential link between multimodal perception and physical flight execution. Moreover, real-world aerial navigation entails multi-stage missions within visually dense environments, necessitating the resolution of intricate distractors and the maintenance of strict sequential logic to fulfill long-horizon objectives.

To address the limitations of VLN for UAVs, we propose Freefly-thinking, a novel end-to-end Vision-Language-Action(VLA)\cite{rt2} model designed to bridge the gap between semantic and control by simultaneously generating explainable Chain-of-Thought(CoT) rationales and continuous flight control vectors. This framework enables autonomous navigation through a dual-head architecture\cite{simlingo} comprising a language adaptor and a waypoint adaptor. Both heads decode features from shared hidden states to ensure synchronized decision-making. Specifically, the language adaptor generates CoT and discrete actions, while the waypoint adaptor predicts continuous relative waypoints and yaw angles. By integrating these processes, our method unifies spatial understanding, control reasoning, and multimodal alignment. We introduce our dataset, built upon OpenFly, to address the critical lack of reasoning annotations in existing UAV VLN benchmarks that merely support direct control mapping. Augmenting the standard vision-instruction-action triplets, we curate the dataset by leveraging a Vision-Language Model (VLM)\cite{internvl, qwenvl} to synthesize explicit CoT rationales for ground-truth controls, while intelligently extracting visual landmarks to demarcate multi-stage trajectory transitions. We adopt a two-stage training paradigm. In the first stage, Supervised Fine-Tuning (SFT)\cite{sft}, both the generated textual rationales and control waypoints are strictly supervised to ensure precise alignment with ground-truth annotations. Subsequently, we introduce a Reinforcement Fine-Tuning (RFT)\cite{rlhf} stage utilizing Group Relative Policy Optimization (GRPO). Guided by carefully crafted reward functions to evaluate and validate the textual outputs, this RFT stage fundamentally enhances the model's reasoning capabilities across the cohesive Image-Instruction-Control logic chain.

Freefly-thinking achieves state-of-the-art navigation performance and introduces explicit reasoning and question-answering (QA) capabilities to the VLN domain. We evaluate the navigation  capabilities of our framework across unseen dataset splits and novel simulated environments. Results demonstrate the superior generalization and leading efficacy compared to existing baselines. 

In summary, our work makes these major contributions:(1) We propose Freefly-thinking, a dual-head VLA framework that bridges the semantic-to-control gap by simultaneously generating explainable CoT rationales and continuous flight control vectors; (2) We curate a comprehensive multi-stage UAV VLN dataset derived from OpenFly, augmenting standard image-instruction-action triplets with explicit CoT reasoning annotations and visual landmarks; (3) We design a two-stage training paradigm featuring SFT and GRPO, enabling the model to achieve state-of-the-art generalization, reasoning, and question-answering performance across unseen environments. \cref{fig:overview} provides an overview of our work and highlights.

\section{Related Work}
\label{sec2}

\subsection{Vision-and-Language Navigation}
\label{sec2:vln}
VLA models and CoT reasoning have revolutionized Embodied AI in domains like robotic manipulation and autonomous driving by enhancing interpretability and complex planning capabilities\cite{vlnr1, alphadrive}. However, extending these paradigms to UAVs introduces profound challenges due to the high-frequency dynamics and unconstrained 6-Degree-of-Freedom motion inherent to 3D aerial environments.

Historically, UAV Vision-and-Language Navigation (VLN) adapted ground-based frameworks to aerial scenarios, pioneered by works like AerialVLN\cite{aeria}. Consequently, UAV VLN in recent years methodologies largely bifurcate into two distinct categories: zero-shot\cite{zeroshot} navigation using frozen LLMs as high-level planners, which often suffer from insufficient continuous spatial perception, and fine-tuned end-to-end models\cite{openfly, autofly, citynav}, which process multimodal inputs to directly output discrete actions or continuous waypoints.

By operating as "black boxes" devoid of explicit reasoning, these existing frameworks suffer from a severe semantic-to-control gap, leading to frequent navigation failures in multi-stage 3D environments. To solve this problem, our Freefly-thinking model introduces a novel dual-head VLA architecture that strictly aligns step-by-step CoT rationales with continuous flight waypoints to unify cognitive planning with precise physical execution.

\subsection{Reinforcement Learning and Reasoning}
\label{sec2:rft}

Reinforcement Fine-Tuning (RFT) performs better for complex tasks\cite{deepseek, deepseekmath}. Specifically, GRPO uses verifiable rewards to elicit explicit Chain-of-Thought reasoning. Despite this, current UAV navigation still relies heavily on pure SFT, which lacks the exploration needed for complex flight dynamics. Our framework solves this by pioneering GRPO-based RFT for UAVs. We design environment-grounded rewards for policy optimization. This strictly aligns textual CoT with continuous physical trajectories.

\section{Methodology}
\label{sec3}
In this section, we introduce our Methodology. \cref{sec3:overview} covers the review of our Freefly-thinking framework, \cref{sec3:vlamodel} details the architecture of the VLA model, \cref{sec3:datset} outlines the dataset construction and data collection pipeline, and finally, \cref{sec3:RL} elaborates the reinforcement learning strategy.
\begin{figure}[tb]
  \centering
  \includegraphics[height=6.5cm]{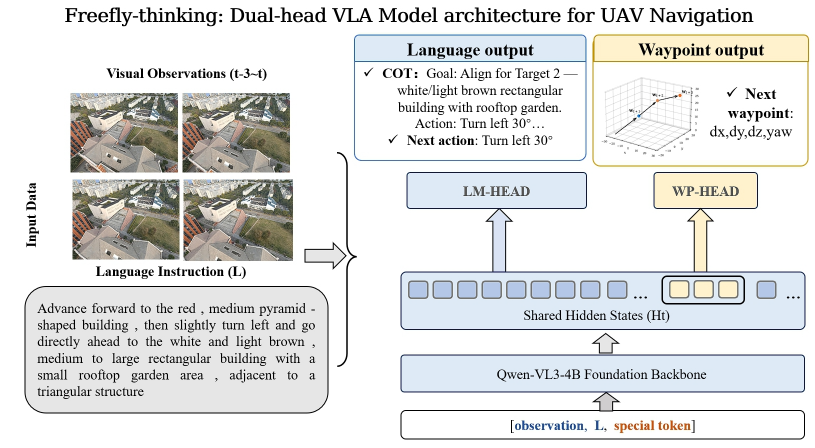}
  \caption{Architecture for Freefly model, which extracts shared multimodal representations from sequential visual observations and language instructions to simultaneously predict explicit Chain-of-Thought reasoning via a Language Model head and continuous 3D spatial waypoints via a Waypoint head}
  \label{fig:model}
\end{figure}

\subsection{Overview}
\label{sec3:overview}

We formulate autonomous navigation as a control policy $\pi$. At each time step $t$, the policy takes the current RGB observation $o_t \in \mathcal{O}$ and the language instruction $L$ as input. Unlike traditional direct-mapping methods, our policy $\pi$ employs a dual-head architecture to generate three distinct output variables: an explicit CoT rationale $r_t \in \mathcal{R}$, a continuous control waypoint $w_t \in \mathcal{W}$ and a discrete control action $a_t \in \mathcal{A}$. Our objective is to derive an optimal policy $\pi^*: (\mathcal{O}, L) \rightarrow (\mathcal{R}, \mathcal{A}, \mathcal{W})$ that generates a collision-free trajectory $\tau = \{w_0, w_1, \dots, w_T\}$. The specific architecture of the model is shown in \cref{fig:model}.

\subsection{VLA model for UAV navigation}
\label{sec3:vlamodel}

UAV navigation in unconstrained 3D environments is complex, as executing long-horizon, multi-stage instructions requires profound spatial awareness. To address this, we introduce explicit CoT reasoning into the navigation framework; more crucially, we couple this CoT generation with continuous waypoint prediction for two critical reasons:

Bridging the Semantic-Kinematic Gap: CoT decomposes ambiguous language instructions into structured, intermediate sub-goals, directly grounding the high-level cognitive planning into explainable and precise low-level flight control vectors.

Anchoring Trajectory Robustness: Step-by-step reasoning acts as a cognitive anchor at each decision juncture, ensuring the waypoints and actions strictly adhere to the correct overarching navigational intent.

Motivated by these insights, we propose Freefly-thinking, a dual-head VLA architecture that integrates reasoning with UAV control. The framework is built upon a Qwen-VL3-4B backbone to extract rich multimodal features. Leveraging the shared hidden states, the model employs two output branches: a Language Model head (LM-head) that autoregressively generates the CoT rationales and discrete action instructions, and an auxiliary continuous waypoint head dedicated to predicting relative 3D waypoints and yaw angles.

\textbf{Vision-Language Model}. We adopt Qwen3-VL as our foundational backbone because its DeepStack architecture inherently provides the exceptional 3D spatial awareness required for complex aerial navigation. The lightweight Qwen3-VL-4B variant is selected to strike the balance between reasoning and the strict computational constraints of  UAV deployment.

\textbf{Dual-head architecture}. We build upon the VLM backbone to process multimodal inputs and extract shared hidden states. Utilizing one and only representation space ensures that both high-level semantic reasoning and low-level spatial planning are grounded in the exact same perceptual context; this multi-task learning paradigm significantly reduces computational overhead while preventing representation misalignment.

During the tokenization and label construction phase, we append three special tokens at the end of the text sequence. The hidden states corresponding to these three tokens are then routed directly into the continuous action head to predict the precise waypoints for the next three future time steps.

Finally, the architecture diverges into two output branches: the LM-head autoregressively generates the CoT rationales and discrete actions, while the Waypoint-head directly outputs the continuous 3D coordinate and yaw vectors.

\subsection{Dataset and Data collection}
\label{sec3:datset}

During trajectory sample construction, each sample is formulated with a global instruction, four sequential visual frames (the current observation alongside three historical frames) and corresponding historical flight actions. To acquire high-quality planning and reasoning data, we use a more powerful large multimodal model Qwen-VL-Plus. By providing this teacher model with prompts containing the flight scenario, we instruct it to generate concise decision-making processes. The generated CoT rationales accurately articulate the current navigation stage, prominent visual landmarks, and logical action planning. Following the quality inspection and the diversified style processing, we obtain a robust dataset of explicit planning rationales.

A primary challenge in UAV navigation datasets is severe class imbalance, as routine "straight" commands dominate the action space. To mitigate this, we introduce a temporal windowing strategy for other critical maneuvers. In details, for the two time steps immediately preceding the execution of a critical action (e.g., turning), we relabel the default "straight" annotations to match the upcoming maneuver. This approach not only augments minority action classes but also aligns closely with real-world flight controls, where initiating a critical operation within a reasonable temporal proximity is both valid and safe.

\begin{figure*}[tb]
  \centering
  \begin{subfigure}{0.32\linewidth}
    \centering
    \includegraphics[width=\linewidth]{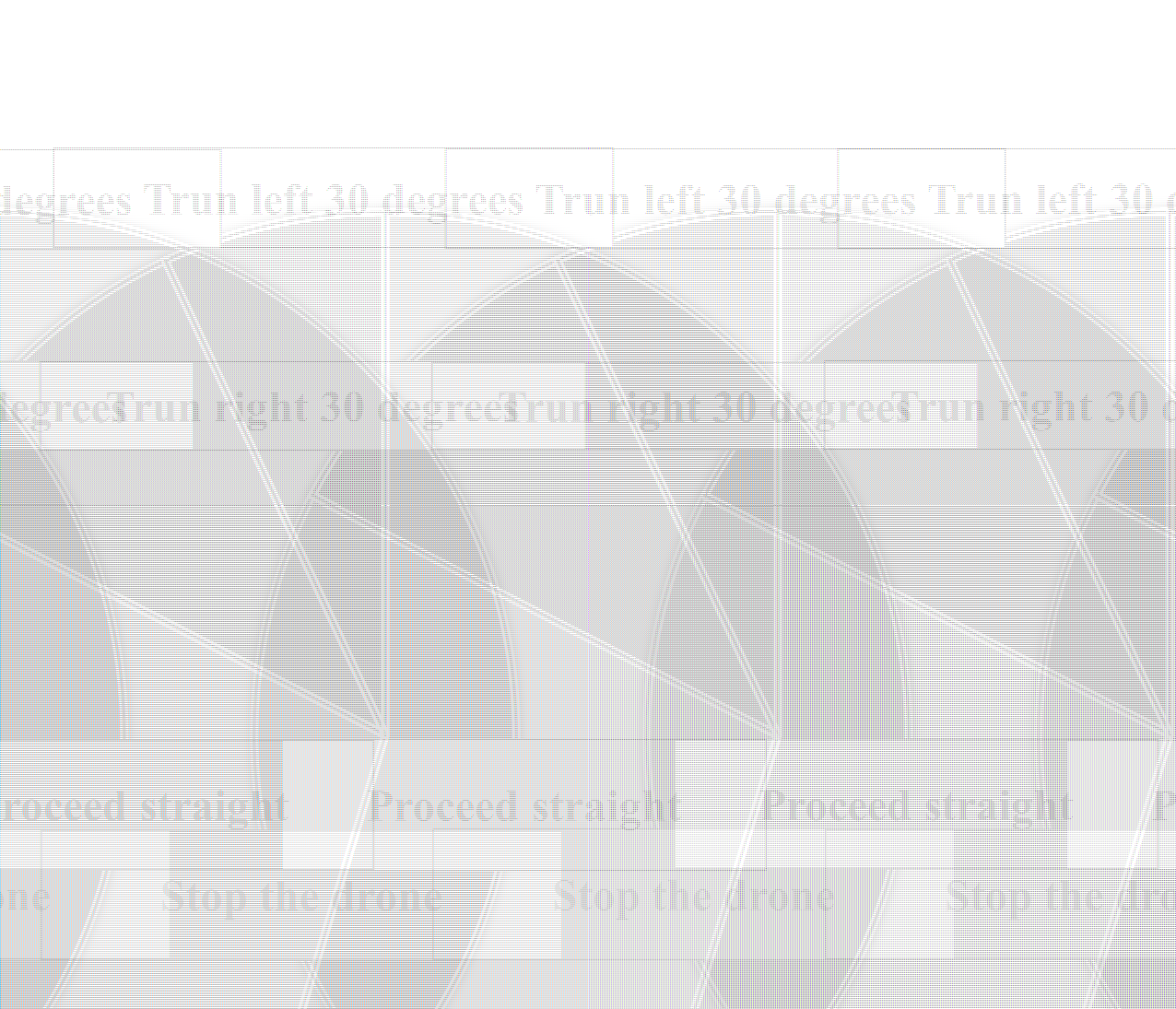}
    \caption{Distribution of primary maneuvers}
    \label{fig:action_dist}
  \end{subfigure}
  \hfill
  \begin{subfigure}{0.32\linewidth}
    \centering
    \includegraphics[width=\linewidth]{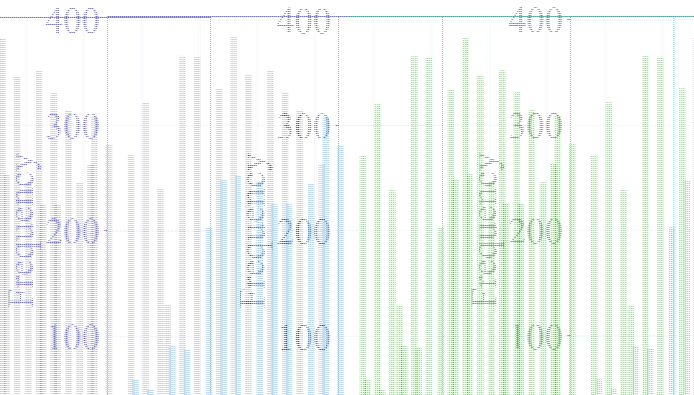}
    \caption{Trajectory length distribution}
    \label{fig:traj_length}
  \end{subfigure}
  \hfill
  \begin{subfigure}{0.32\linewidth}
    \centering
    \includegraphics[width=\linewidth]{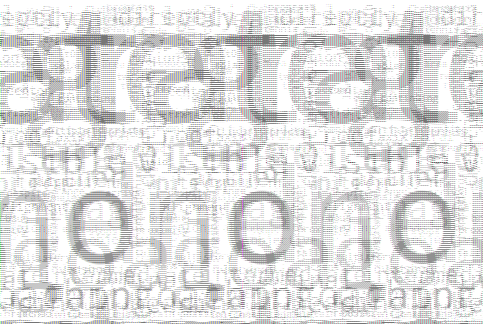}
    \caption{Word cloud of CoT rationales}
    \label{fig:cot_cloud}
  \end{subfigure}
  
  \caption{Statistical overview of our constructed UAV navigation dataset. (a) The sample size distribution of primary flight maneuvers (e.g., straight, turn). (b) The distribution of physical trajectory lengths. (c) A word cloud visualization highlighting the most frequent terms in the generated Chain-of-Thought (CoT) rationales.}
  \label{fig:dataset_statistics}
\end{figure*}

Overall, the final dataset comprises 6820 short trajectories, yielding a total of 101220 images, with an average of 2.89 critical operations 81 meters length per trajectory. We partition this dataset into a 95$\%$ training set and a 5$\%$ validation set, while an unseen test set consisting of 192 trajectories is reserved for evaluation. During the training process, the entirety of the training set is utilized for SFT. In RFT stage, we construct a high-quality subset using the CLIP model by extracting the top 10,000 samples with the highest similarity scores. This RFT subset maintains a 4:6 ratio between straight flights and critical maneuvers. The visual characteristics of the dataset are shown in \cref{fig:dataset_statistics}

\subsection{Two-Stage Training}
\label{sec3:RL}

We employ a two-stage training paradigm consisting of SFT and RFT. SFT utilizes dense supervision to clone expert behaviors, while RFT optimizes Instruction-Image-Action alignment through exploration-driven learning.

\textbf{SFT stage}. We optimize the dual-head model to mimic expert trajectories. Given the visual observation $o_t$ and instruction $L$, the model jointly predicts a textual sequence $y_t$ (containing the CoT rationale and discrete action) and a continuous $n$-step waypoint trajectory $\hat{\mathbf{w}}_{t+1:t+n}$ ($n=3$). For the LM-head, we minimize the standard autoregressive cross-entropy loss over the ground-truth text tokens $y^*=\{y_1^*,y_2^*,\dots,y_M^*\}$:

\begin{align}
\mathcal{L}_{\text{LM}}=-\sum_{j=1}^{M}\log P(y_j^*|y_{1:j-1}^*,o_t,L)
\end{align}

Concurrently, for the Waypoint-head, we apply an L1 regression loss between the predicted spatial waypoints $\hat{\mathbf{w}}_{t+k}$ and the expert ground-truth $\mathbf{w}^*_{t+k}$:

\begin{align}
\mathcal{L}_{\text{WP}}=\sum_{k=1}^{n}\|\hat{\mathbf{w}}_{t+k}-\mathbf{w}^*_{t+k}\|_1
\end{align}

The overall SFT loss is computed as the weighted sum of both branches to jointly learn compositional reasoning and spatial kinematics:

\begin{align}
\mathcal{L}_{\text{SFT}}=\mathcal{L}_{\text{LM}}+\lambda\mathcal{L}_{\text{WP}}
\end{align}

where $\lambda$ is a learnable parameter balancing the textual reasoning and continuous physical control objectives.

\textbf{RFT stage}. We select GRPO as our reinforcement learning algorithm for two reasons. First, famous models like DeepSeek-R1 prove that GRPO provides superior training stability and computational efficiency. Second, its group-based relative optimization perfectly suits UAV navigation, since spatial planning entails multiple flight action choices to next targets in a traj navigation.

To enforce Image-Instruction-Action alignment, we design four verifiable reward functions to guide the GRPO exploration.

Format Reward: This reward strictly regularizes the model's output structure. It ensures the cognitive reasoning and final actions are properly enclosed in predefined XML tags for downstream execution.

Action Correctness Reward: This metric evaluates whether the predicted discrete action perfectly matches the expert ground truth. It heavily rewards the model when the generated CoT reasoning explicitly leads to the correct physical maneuver. This ensures that the flight action is driven by strict logical deduction rather than random guessing.

Grounding Correctness Reward: We utilize an external vision-language reranker to verify visual-textual semantic alignment. It accurately scores whether the generated rationales explicitly observe the correct visual landmarks and navigation phases. This verification minimizes visual cognitive errors, providing a more solid CoT foundation.

Length Penalty Reward: We encourage the model to generate slightly longer rationales than the expert samples to foster deeper reasoning. However, we strictly penalize excessively verbose text during exploration. This forces concise reasoning and guarantees low inference latency for real-time flight.

These rewards are aggregated to compute the GRPO loss for parameter optimization. Notably, only the language generation parameters are updated during this stage.

\section{Experiments}
\label{sec4}

\subsection{Experimental details}
\label{sec4:detail}
This section presents the experimental details, evaluation performance, and ablation experiments of our proposed approach.

\textbf{SFT training details}. Our model is initialized from the Qwen3-VL-4B-Instruct backbone and optimized for one epoch using a cosine learning rate scheduler with a peak learning rate of $2 \times 10^{-5}$, a 10$\%$ warmup ratio, and a weight decay of 0.01. To ensure computational efficiency and memory optimization during training, we employ bfloat16 precision alongside FlashAttention-2, maintaining a per-device batch size of 6 with 8 gradient accumulation steps.

\textbf{RFT training details}. For the RFT stage, we adopt a reduced peak learning rate of $2 \times 10^{-6}$ and apply a KL divergence penalty coefficient ($\beta$) of 0.001. To encourage diverse trajectory exploration, we sample $G=4$ rollouts per prompt during training, utilizing a stochastic decoding strategy configured with a temperature of 0.9, a top-$p$ of 0.9, and a top-$k$ of 40. The hardware setup maintains the identical bfloat16, while adjusting the per-device batch size to 8 with 8 gradient accumulation steps to stabilize the reinforcement learning updates.

\textbf{Evaluation Metrics}. We adopt four key metrics to evaluate our model. Success Rate(SR) measures the percentage of flight episodes where the UAV's final predicted position is bounded within a 20-meter radius of the intended destination. Navigation Error(NE) quantifies the final spatial distance between the predicted trajectory endpoint and the actual ground-truth target. Average Displacement Error(ADE) computes the mean spatial deviation between the predicted sequence of waypoints and the expert trajectory across all $n$ time steps. 
It is formally defined as:

\begin{align}
\text{ADE}=\frac{1}{n}\sum_{k=1}^{n}\|\hat{\mathbf{w}}_{t+k}-\mathbf{w}^*_{t+k}\|_2
\end{align}

Sample-wise Action Prediction Accuracy evaluates the classification correctness of the model's discrete linguistic maneuver predictions (e.g., straight, turn left) against the expert annotations at each individual time step.

\subsection{Evaluation performence}
\label{sec4:unseen}

To ensure a fair and rigorous comparison, all baseline methods are evaluated under the exact same training dataset and evaluation protocol. The quantitative results on the unseen test set are summarized in \cref{tab:evaluation_unseen}. Our Freefly-thinking model(best sft version) consistently outperforms established baselines, achieving the lowest Navigation Error (28.0m) and Average Displacement Error (2.3m) alongside a state-of-the-art Success Rate of 13.1$\%$. The performance gap stems directly from inherent architectural differences: AerialVLN relies on a traditional model that severely struggles with complex 3D spatial perception, while the stronger OpenFly baseline lacks intermediate logical planning by directly predicting flight maneuvers. In contrast, our dual-head generates step-by-step CoT rationales prior to physical execution, effectively enabling the agent to reason before acting and significantly mitigating continuous trajectory deviations.

\begin{table}[tb]
  \renewcommand{\arraystretch}{1.2}
  \setlength{\tabcolsep}{4mm}
  
  \caption{Evaluation results on the unseen test set. 
    NE and ADE are measured in meters (lower is better), while SR and Act. Acc are presented as percentages (higher is better). 
  }
  \label{tab:evaluation_unseen}
  \centering
  \begin{tabular}{@{}lcccc@{}}
    \toprule
    Method & NE$\downarrow$ & SR$\uparrow$ & ADE$\downarrow$ \\
    \midrule
    AerialVLN & 45.9 & 4.3\% & 3.7 \\
    OpenFly & 32.7 & 11.3\% & 2.6 \\
    Ours & \textbf{28.0} & \textbf{13.1\%} & \textbf{2.3} \\
  \bottomrule
  \end{tabular}
\end{table}

\subsection{Ablation experiments}
\label{sec4:ablation}

\cref{tab:evaluation_ab} details our ablation experiments across three configurations. The results validate our architectural design. The <SFT dual head> model strictly outperforms the <SFT w/o thinking> baseline. Specifically, Waypoint-head Success Rate (SR) improves from 11.0$\%$ to 13.1$\%$. Its Navigation Error (NE) also decreases from 31.8 to 28.0. This confirms that explicit reasoning directly enhances continuous physical execution. Furthermore, applying Reinforcement Fine-Tuning (RFT) maximizes cognitive performance. Under RFT, the LM-head achieves the best reasoning metrics, reaching a peak SR of 30.4$\%$ and an Action Accuracy of 84.5$\%$. RFT did not optimize the Waypoint-head. It traded a decrease in waypoint precision for superior logical planning capabilities. This outcome aligns perfectly with our expectations.

\begin{table}[tb]
  \renewcommand{\arraystretch}{1.1}
  \setlength{\tabcolsep}{2.5mm}
  
  \caption{Evaluation results for ablation experiments.
  }
  \label{tab:evaluation_ab}
  \centering
  \begin{tabular}{@{}l cccc ccc@{}}
    \toprule
    Method & \multicolumn{4}{c}{LM-head} & \multicolumn{3}{c}{Waypoint-head} \\
    & NE$\downarrow$ & SR$\uparrow$ & ADE$\downarrow$ & Act. Acc$\uparrow$ & NE$\downarrow$ & SR$\uparrow$ & ADE$\downarrow$ \\
    \midrule
    RFT & 16.5 & 30.4\% & 1.4 & 84.5\% & 35.1 & 9.6\% & 3.2 \\
    SFT w/o thinking  & 19.1 & 27.2\% & 1.5 & 83.0\% & 31.8 & 11.0\% & 2.5 \\
    SFT dual head     & 18.7 & 27.6\% & 1.5 & 83.1\% & 28.0 & 13.1\% & 2.3 \\
  \bottomrule
  \end{tabular}
\end{table}

\section{Conclusion}

Current UAV Vision-and-Language Navigation (VLN) models typically act as black boxes without explicit reasoning. To overcome this limitation, we propose Freefly-thinking, a novel dual-head Vision-Language-Action (VLA) architecture designed for aerial navigation. Our model simultaneously predicts explicit step-by-step Chain-of-Thought (CoT) rationales and continuous 3D spatial waypoints. We optimize this framework through Supervised Fine-Tuning (SFT) for foundational control alignment and Reinforcement Fine-Tuning (RFT) for enhanced logical planning. Extensive experiments demonstrate that our approach significantly outperforms established baselines in both spatial trajectory precision and overall navigation success rate.

\bibliographystyle{splncs04}
\bibliography{main}
\end{document}